\documentclass{article}
\usepackage{spconf,amsmath,graphicx,hyperref}
\usepackage{microtype}
\usepackage{amsfonts}
\usepackage{babel}
\usepackage{booktabs}
\usepackage{graphicx}
\usepackage{multirow}
\usepackage{array}
\usepackage{xcolor}
\usepackage{adjustbox} 
\usepackage{enumitem}
\usepackage{hyperref}

\definecolor{darkgreen}{RGB}{0,128,0}
\definecolor{darkyellow}{RGB}{204, 153, 0}

\title{RIS-FUSION: Rethinking Text-Driven Infrared and Visible Image Fusion from the Perspective of Referring Image Segmentation}

\name{
  Siju Ma$^{1}$,
  Changsiyu Gong$^{1}$,
  Xiaofeng Fan$^{1}$,
  Yong Ma$^{1}$,
  Chengjie Jiang$^{2*}$\thanks{*Corresponding author}
}
\vspace{-1pt} 

\address{
  $^{1}$ Central University of Finance and Economics, Beijing, China \\
  $^{2}$ Tsinghua University, Shenzhen, China
}
\vspace{-5pt} 

\begin{document}
%
\maketitle
\begin{abstract}
Text-driven infrared and visible image fusion has gained attention for enabling natural language to guide the fusion process. However, existing methods lack a goal-aligned task to supervise and evaluate how effectively the input text contributes to the fusion outcome. We observe that referring image segmentation (RIS) and text-driven fusion share a common objective: highlighting the object referred to by the text. Motivated by this, we propose \textbf{RIS-FUSION}, a cascaded framework that unifies fusion and RIS through joint optimization. At its core is the \textit{LangGatedFusion} module, which injects textual features into the fusion backbone to enhance semantic alignment. To support multimodal referring image segmentation task, we introduce \textit{MM-RIS}, a large-scale benchmark with 12.5k training and 3.5k testing triplets, each consisting of an infrared-visible image pair, a segmentation mask, and a referring expression. Extensive experiments show that \textbf{RIS-FUSION} achieves state-of-the-art performance, outperforming existing methods by over 11\% in mIoU. Code and dataset will be released at \href{https://github.com/SijuMa2003/RIS-FUSION}{https://github.com/SijuMa2003/RIS-FUSION}.
\end{abstract}

\section{Introduction}
\label{sec:intro}

Infrared and Visible Image Fusion(IVIF) is a fundamental task in digital image processing~\cite{PSFusion,maefuse}. Single–modal images capture only a partial representation of a scene, whereas multimodal imagery provides a more comprehensive view. Specifically, visible (VIS) images are effective in capturing fine-grained color and texture cues, whereas infrared (IR) images, leveraging thermal radiation, offer precise structural information under low illumination, fog, or other adverse conditions. Combining them through IVIF allows both modalities to contribute to a single, high-quality result. Effective IVIF models should not only ensure good perceptual quality but also retain semantic information for downstream tasks, making them valuable in applications such as surveillance and autonomous driving~\cite{hsfusion,rfnnest,e2emfd,rethinking}.

Recent IVIF works have explored the potential of \emph{text-driven} image fusion, where natural-language prompts guide the fusion process~\cite{textfusion,terf,textdifuse,textif,omnifuse,ditfuse}. These methods aim to generate controllable fused images that highlight user-specified content. For instance, TextFusion~\cite{textfusion} and TeRF~\cite{terf} follow a two-stage paradigm: they first produce a general fused image, then refine it using text-guided enhancement modules. This enables post-hoc control for visualization or downstream task adaptation. However, once the fusion result is fixed, the refinement stage cannot reselect modality-specific information, limiting its flexibility. In contrast, Text-DiFuse~\cite{textdifuse} and Text-IF~\cite{textif} integrate text-control modules within the fusion network to locally suppress degraded regions and enhance target objects. However, they lack a principled evaluation method to quantify whether the fused output truly benefits from the input text. OmniFusion~\cite{omnifuse} takes a further step by supporting flexible semantic control through joint optimization with a semantic segmentation task. It validates the impact of text guidance by comparing segmentation performance with and without text control. However, semantic segmentation predicts all classes in a scene, while text-driven fusion usually targets a single object described by the prompt. This misalignment makes semantic segmentation a suboptimal objective for evaluating or guiding text-driven IVIF.

To address this gap, we rethink text-driven IVIF from the perspective of \emph{referring image segmentation} (RIS). RIS defines segmentation as a binary task that highlights the object specified by a natural-language expression. This property aligns well with the goal of text-driven fusion. As a result, RIS provides a more suitable and task-aligned supervision signal for guiding and evaluating text-conditioned fusion.

\begin{figure*}[!t]
  \centering
  \includegraphics[width=\textwidth]{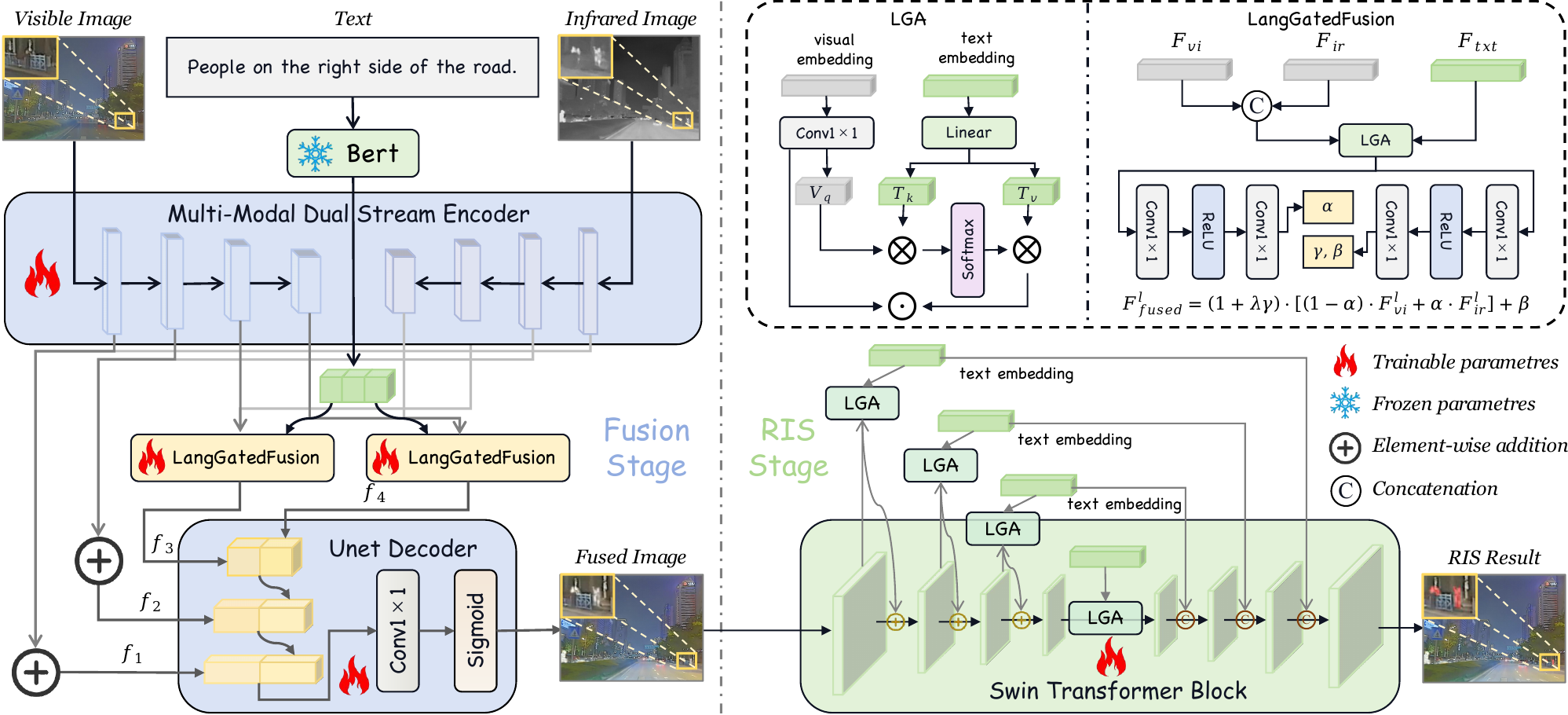} 
  \caption{
    The overall architecture of the proposed \textbf{RIS-FUSION} framework.
    }
  \label{fig:model_arch}
\end{figure*}

Building on this insight, we propose \textbf{RIS-FUSION}, a cascaded architecture that unifies fusion and RIS. As shown in Fig.~\ref{fig:model_arch}, we design a novel \textit{LangGatedFusion} module that encodes the expression with BERT~\cite{bert} and injects it into the fusion encoder via cross-attention. The fused image and text features are then passed to a Swin Transformer–based~\cite{swintransformer} RIS head. Besides, the RIS loss is back-propagated to the fusion stage, aligning the fusion process with textual referents and downstream objectives. To support this \textit{M}ulti\textit{M}odal \textit{R}eferring \textit{I}mage \textit{S}egmentation task, we construct the \textit{MM-RIS} dataset by annotating $M^{3}FD$~\cite{TarDaL} and $MSRS$~\cite{PIAFusion} with referring expressions and fine-grained masks. \textit{MM-RIS} contains 12.5k training and 3.5k testing triplets of IR–VIS images with associated expressions and annotations. Experiments on \textit{MM-RIS} demonstrate that \textbf{RIS-FUSION} generates clearer fused images and achieves state–of–the–art RIS performance.

The main contributions can be summarized as follows:

\begin{itemize}[itemsep=0pt, topsep=0pt, parsep=0pt, partopsep=0pt]
    \item We introduce a RIS–driven perspective for evaluating and optimizing text–driven IVIF, providing a more precise measure of language–conditioned fusion.
    \item We propose \textbf{RIS-FUSION}, a cascaded architecture that connects IVIF with RIS. The \textit{LangGatedFusion} module injects textual information directly into the fusion process, and the entire framework is jointly optimized to ensure semantic consistency.
    \item We construct the \textit{MM-RIS} dataset, containing 12.5k training and 3.5k testing IR–VIS pairs with fine-grained segmentation masks and expressions. Our method achieves state-of-the-art performance on this benchmark, outperforming existing approaches by over 11\% in RIS mIoU, demonstrating significant improvement.
\end{itemize}

\section{Methods}
\label{sec:methods}
\subsection{Overall Architecture}

As shown in Fig.~\ref{fig:model_arch}, the proposed \textbf{RIS-FUSION} framework adopts a cascaded two-stage pipeline, consisting of a \emph{Fusion Stage} followed by a \emph{Referring Image Segmentation (RIS) Stage}. The inputs to the system include a visible image $I_{\text{vi}} \in \mathbb{R}^{3 \times H \times W}$, an infrared image $I_{\text{ir}} \in \mathbb{R}^{1 \times H \times W}$, and a natural-language expression $T$ that describes the target object using attributes or spatial cues.

In the preprocessing step, the visible image $I_{\text{vi}}$ is converted to YCbCr color space, and only its luminance channel $Y_{\text{vi}} \in \mathbb{R}^{1 \times H \times W}$ is used for fusion. The grayscale infrared image is treated as $Y_{\text{ir}}$. The remaining chrominance channels (Cb, Cr) are preserved and later recombined with the fused luminance to reconstruct the RGB output.

In the Fusion Stage, $Y_{\text{vi}}$ and $Y_{\text{ir}}$ are processed by a dual-stream pyramid encoder to extract modality-specific features $F_{\text{vi}}^l$ and $F_{\text{ir}}^l$ at level $l$. The two encoders do not share weights and operate at four resolution levels. Features from the first two scales are fused via element-wise addition to preserve low-level structural and textural cues. At the deeper two scales, fusion is guided by the text embedding $F_{\text{txt}} \in \mathbb{R}^{N \times d}$, obtained from a frozen BERT~\cite{bert} encoder, and performed through our proposed LangGatedFusion modules. The resulting multi-scale features are decoded via a U-Net decoder with skip connections, yielding a fused luminance image $Y_{\text{fuse}}$, which is then recombined with the original Cb and Cr channels to reconstruct the final RGB fused image $\hat{I}_{\text{fuse}}$.

In the RIS Stage, the fused image $\hat{I}_{\text{fuse}}$ is fed into a Swin Transformer–based encoder–decoder structure. The same text embedding $F_{\text{txt}}$ is injected into multiple encoder layers via Language-Guided Attention (LGA), and also concatenated with decoder features to enhance cross-modal alignment. The network outputs a binary segmentation mask corresponding to the prompt $T$, which is overlaid as a red semi-transparent mask on the fused image to form the final visual output.

\subsection{Language-Gated Fusion Module}

Given modality-specific features $F_{\text{vi}}^l, F_{\text{ir}}^l \in \mathbb{R}^{C \times H \times W}$, and a text embedding $F_{\text{txt}}$, the LangGatedFusion module performs text-guided multimodal fusion as follows.

We first concatenate the visual features from the visible and infrared branches, then apply a \emph{Language-Guided Attention} (LGA) mechanism with the text features to obtain a text-conditioned context map:
\begin{equation}
F_{\text{ctx}} = \text{LGA}\big(\text{Concat}(F_{\text{vi}}^l,\ F_{\text{ir}}^l),\ F_{\text{txt}}\big) \in \mathbb{R}^{d \times H \times W}.
\end{equation}

In the LGA, the concatenated visual features are reshaped as pixel queries $Q \in \mathbb{R}^{HW \times d}$, and the text embeddings are projected into keys and values $K, V \in \mathbb{R}^{N \times d}$, where $N$ is the number of tokens in the referring expression. The attention output is then computed as:
\begin{equation}
F_{\text{ctx}} = \text{Reshape}\big(\text{Softmax}(QK^\top / \sqrt{d})V\big) \in \mathbb{R}^{d \times H \times W}.
\end{equation}

The context $F_{\text{ctx}}$ is then used to predict the soft spatial gate $\alpha$ together with FiLM modulation parameters $\gamma$ and $\beta$:
\begin{align}
    \alpha &= \sigma(\text{Conv}_\alpha(F_{\text{ctx}})) \in [0,1]^{1 \times H \times W}, \\
    \gamma, \beta &= \text{Conv}_{\text{film}}(F_{\text{ctx}}) \in \mathbb{R}^{1 \times H \times W}.
\end{align}

The final fused output is computed as:
\begin{equation}
    F_{\text{fused}}^l = (1 + \lambda \cdot \gamma) \cdot \left[(1 - \alpha) \cdot F_{\text{vi}}^l + \alpha \cdot F_{\text{ir}}^l \right] + \beta,
\end{equation}
where $\lambda$ is a small scaling factor controlling the strength of FiLM~\cite{film} modulation. This formulation allows dynamic, pixel-level control over modality contribution based on textual semantics, enabling fine-grained, object-aware fusion.

\subsection{Loss Function}

The total loss consists of a segmentation and a fusion loss:
\begin{equation}
    \mathcal{L}_{\text{total}} = \mathcal{L}_{\text{seg}} + \lambda_{\text{fuse}} \cdot \mathcal{L}_{\text{fuse}}.
\end{equation}

\textbf{Segmentation loss.}  
We adopt a dual-class Dice loss to optimize the binary mask output. Let $P \in [0,1]^{H \times W}$ be the predicted mask and $G \in \{0,1\}^{H \times W}$ the ground truth:
\begin{equation}
    \mathcal{L}_{\text{seg}} = 1 - \frac{2 \sum P G + \varepsilon}{\sum P + \sum G + \varepsilon}.
\end{equation}

\textbf{Fusion loss.}  
Let $Y_{\text{fuse}}, Y_{\text{vi}}, Y_{\text{ir}} \in [0,1]^{H \times W}$ denote the luminance ($Y$) channels of the fused, visible, and infrared images, respectively. The fusion loss is defined as:
\begin{equation}
\begin{aligned}
\mathcal{L}_{\text{fuse}} =\;&
\omega_{\text{ssim}}^{\text{vi}} \left(1 - \mathrm{SSIM}(Y_{\text{fuse}}, Y_{\text{vi}})\right) \\
&+ \omega_{\text{mse}}^{\text{vi}} \left\|Y_{\text{fuse}} - Y_{\text{vi}}\right\|_2^2
+ \omega_{\text{mse}}^{\text{ir}} \left\|Y_{\text{fuse}} - Y_{\text{ir}}\right\|_2^2 \\
&+ \omega_{\text{sobel}}^{\text{ir}} \left\|\nabla Y_{\text{fuse}} - \nabla Y_{\text{ir}}\right\|_1 \\
&+ \omega_{\text{grad}} \left(
\left\|\nabla_x Y_{\text{fuse}} - \max(\nabla_x Y_{\text{vi}}, \nabla_x Y_{\text{ir}})\right\|_1 \right. \\
&\quad\quad\left. + \left\|\nabla_y Y_{\text{fuse}} - \max(\nabla_y Y_{\text{vi}}, \nabla_y Y_{\text{ir}})\right\|_1
\right),
\end{aligned}
\end{equation}
here, $\omega_{\text{ssim}}^{\text{vi}}=0.5$, $\omega_{\text{mse}}^{\text{vi}}=0.5$, and $\omega_{\text{mse}}^{\text{ir}}=2.0$ are the weights for the SSIM and MSE losses. $\omega_{\text{sobel}}^{\text{ir}}=1.0$ is the weight for the Sobel edge loss, and $\omega_{\text{grad}}=1.0$ is the weight for the gradient consistency loss.

\textbf{Joint optimization.}  
Fusion and segmentation branches are trained jointly. The fused image $\hat{I}_{\text{fuse}}$ is fed into the RIS network \textit{without gradient detachment}, allowing the segmentation loss $\mathcal{L}_{\text{seg}}$ to back-propagate into the fusion module. This design encourages mutual promotion: segmentation guides semantic fidelity of fusion, while fusion optimizes structural quality for better downstream segmentation.

\section{Experiments}
\label{sec:implementation_details}
\begin{table*}[htbp]
\centering
\renewcommand{\arraystretch}{1.0}
\caption{Quantitative Comparison on \textit{MM-RIS} Test Set.}
\label{tab:ris_comparison}
\begin{adjustbox}{width=\linewidth}
\begin{tabular}{lccccccc}
\toprule
Method & Pub. & P@0.5 & P@0.6 & P@0.7 & P@0.8 & P@0.9 & mIoU \\
\midrule
VIS+RIS            & -         & 33.54 & 25.52 & 17.68 & 10.48 & 4.31 & 34.89  \\
TeRF+RIS$_{w/o}$           & MM'24     & 45.32     & 37.05     & 27.14     & 17.63     & 7.34    & 45.33      \\
TextFusion+RIS$_{w/o}$     & IF'25     & 42.88 & 34.31 & 25.63 & 16.45  & 7.49 & 42.77  \\
Text-DiFuse+RIS$_{w/o}$    & NeurIPS'24 & 44.15     & 35.86     & 27.02     & 17.88     & 7.34    & 45.02      \\
Text-IF+RIS$_{w/o}$        & CVPR'24   & 46.77     & 37.01     & 27.52     & 18.35     & 8.00    & 46.39  \\
OmniFuse+RIS$_{w/o}$       & TPAMI'25  & 49.98 & 41.25 & 30.20 & 20.56 & 8.00 & 47.16   \\
\midrule
TeRF+RIS$_w$           & MM'24     & 47.87     & 38.19     & 27.88     & 18.60     & 8.06    & 46.12      \\
TextFusion+RIS$_w$     & IF'25     & 43.30 & 37.57 & 27.00 & 17.81  & 7.99 & 45.48  \\
Text-DiFuse+RIS$_w$    & NeurIPS'24 & 46.35     & 37.81     & 28.26     & 18.03     & 8.34    & 47.02  \\
Text-IF+RIS$_w$        & CVPR'24   & 48.87     & 39.98     & 29.10     & 19.24     & 9.00    & 47.68   \\
OmniFuse+RIS$_w$       & TPAMI'25  & 50.42 & 42.27 & 32.95 & 22.01  & 8.75 & 47.71   \\
\midrule
\textbf{RIS-FUSION$_{w/o}$} & - & 
\underline{52.18} \textcolor{darkyellow}{(+3\%)} &
\underline{44.36} \textcolor{darkyellow}{(+5\%)} &
\underline{34.90} \textcolor{darkyellow}{(+6\%)} &
\underline{24.28} \textcolor{darkyellow}{(+10\%)} &
\underline{10.88}  \textcolor{darkyellow}{(+21\%)} &
\underline{48.91} \textcolor{darkyellow}{(+3\%)} \\
\textbf{RIS-FUSION$_w$} & - & 
\textbf{58.81} \textcolor{darkgreen}{(+17\%)} &
\textbf{50.23} \textcolor{darkgreen}{(+19\%)} &
\textbf{40.17} \textcolor{darkgreen}{(+22\%)} &
\textbf{28.41} \textcolor{darkgreen}{(+29\%)} &
\textbf{13.23}  \textcolor{darkgreen}{(+47\%)} &
\textbf{53.00} \textcolor{darkgreen}{(+11\%)} \\
\bottomrule
\end{tabular}
\end{adjustbox}
\vspace{-8pt}
\end{table*}
\subsection{Implementation Details}
All experiments are conducted on a single Ascend snt9b3 card. The training process uses the AdamW optimizer, with separate learning rates for segmentation and fusion modules: $5\times10^{-5}$ for segmentation and $1\times10^{-4}$ for pre-fusion, both with a weight decay of $1\times10^{-2}$. The model is trained for 10 epochs with a batch size of 16. All input images are resized to $480\times480$ during training. The \textit{MM-RIS} dataset is used, with a 4:1 split for training and validation.

\begin{figure}[t] 
    \centering
    \vspace{-5pt}
    \includegraphics[width=\columnwidth]{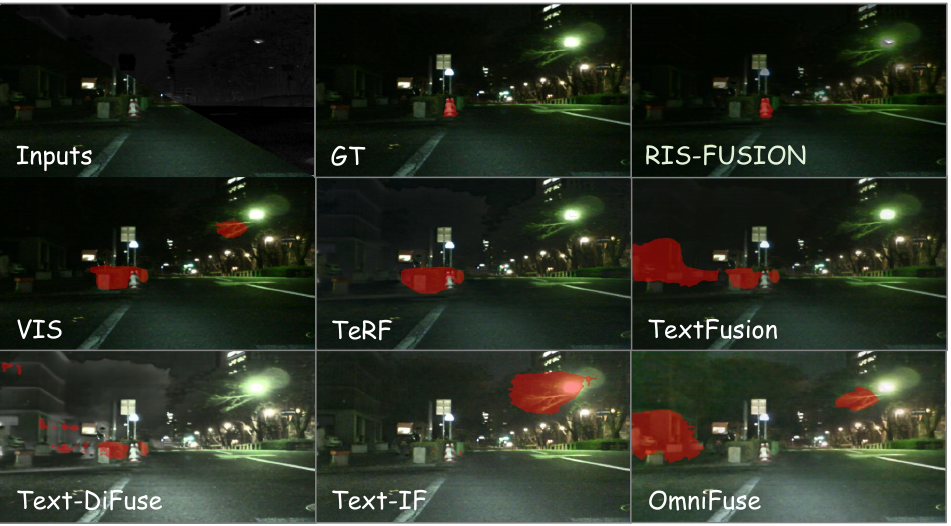}
    \vspace{-5pt}
    \caption{Qualitative comparison of the multimodal referring image segmentation task. The referring text is ``two color cones on the left side of the road''.}
    \label{fig:qualitative_result}
\end{figure}

\subsection{\textit{MM-RIS} Dataset}
To facilitate a unified benchmark for multimodal referring image segmentation, we construct the \textit{MM-RIS} Dataset by building upon two widely-used visible and infrared image fusion datasets: $MSRS$~\cite{PIAFusion} and $M^{3}FD$~\cite{TarDaL}. Specifically, we first categorize all semantic segmentation masks from the original datasets according to their class labels. For each semantic category, we then separate all disconnected mask regions and treat each of them as an individual candidate target.

Human annotators are then presented with these candidate regions in the fused images and asked to select one or more regions and provide a natural-language expression that uniquely refers to them. These expressions may describe the object using attributes (e.g., color, size, shape) or spatial relationships (e.g., “the leftmost building,” “tree beside the car”). This results in a total of 12.5k samples for training and 3.5k samples for testing, each consisting of a visible image, an infrared image, a fused image, a referring expression, and a binary segmentation mask of the referred region.

To ensure a fair comparison, we use mean fusion results as the inputs to train the Swin Transformer-based LAVT~\cite{lavt} segmentation backbone with Dice loss. This setup adapt the segmentation model to the IVIF domain. Besides, all compared baselines are \textit{text-driven} IVIF methods, where the input text is formed by prepending “please make xxx clear” to the RIS-style referring expression.

\subsection{Qualitative Results}
Fig.~\ref{fig:qualitative_result} presents the qualitative comparison results. Here, \textit{Inputs} denotes the paired infrared and visible images, while \textit{VIS} refers to the segmentation output of LAVT with only the visible modality as input. It can be observed that our proposed method effectively suppresses environmental factors such as over-exposure and highlights the target objects (color cones), thereby achieving precise referring image segmentation. In contrast, other approaches tend to focus on other salient objects, such as bollards, pedestrians, or street lamps.

\subsection{Quantitative Results}
Tab.~\ref{tab:ris_comparison} reports the quantitative comparison. Subscripts \textit{w/o} and \textit{w} denote testing the fusion module \emph{without} and \emph{with} the referring text, respectively. Our method already attains state-of-the-art performance in the \textit{w/o} setting, which we attribute to the proposed joint optimization strategy between fusion and segmentation. When enabling text-driven fusion (\textit{w}), our LangGatedFusion further injects semantic guidance from the referring text, yielding consistent gains across all thresholds and up to 11\% relative improvement in mIoU. In addition, we observe a noticeable performance drop using only the visible image, highlighting the necessity of leveraging multimodal information for referring image segmentation.

\subsection{Ablaion Study}
We conduct ablation experiments on the LangGatedFusion module and the joint optimization strategy, as reported in Tab.~\ref{tab:ablation}. Introducing joint optimization between IVIF and RIS leads to substantial performance improvements, indicating a strong alignment between the objectives of RIS and text-driven fusion. On top of this, incorporating the LangGatedFusion module, which injects textual semantics into the fusion backbone, yields further gains in both mIoU and precision.
\vspace{-6pt}
\begin{table}[htbp]
    \centering
    \caption{Ablation study on the impact of LangGatedFusion module and joint optimization.}
    \small  
    \renewcommand{\arraystretch}{0.95}
    \begin{tabular*}{\columnwidth}{@{\extracolsep{\fill}}rcc|cc}
        \toprule
          & LangGatedFusion & Joint Opt. & mIoU & P@0.5 \\
        \midrule
        1 & $\times$    & $\times$    & 45.47   & 46.99 \\
        2 & $\times$    & \checkmark  & 51.43   & 54.20 \\
        3 & \checkmark  & $\times$    & 47.88   & 51.02 \\
        4 & \checkmark  & \checkmark  & \textbf{53.00}   & \textbf{58.81} \\
        \bottomrule
    \end{tabular*}
    \label{tab:ablation}
\end{table}
\vspace{-8pt}

\section{Conclusion}
\label{sec:conclusion}
In this work, we rethink the objective of text-driven image fusion and identify referring image segmentation (RIS) as a goal-aligned task for supervision and evaluation. Based on this insight, we propose \textbf{RIS-FUSION}, a cascaded framework with a novel \textit{LangGatedFusion} module and joint optimization scheme. We also introduce \textit{MM-RIS}, a large-scale dataset with 16k IR-VIS pairs, masks, and referring expressions. Our method achieves state-of-the-art results with over 11\% mIoU gain. This work provides a new paradigm for evaluating and guiding text-driven image fusion systems.

\clearpage
\bibliographystyle{IEEEbib}
\bibliography{main,refs}

\end{document}